\documentclass[letterpaper, 10 pt, conference]{ieeeconf} 

\IEEEoverridecommandlockouts
\usepackage{cite}
\usepackage{graphicx}
\usepackage{soul}
\usepackage{color}
\graphicspath{{figs/}}
\usepackage{amsmath}
\usepackage{caption}
\usepackage{hyperref}
\begin{document}
\bstctlcite{IEEEexample:BSTcontrol}

\title{\LARGE \bf A Sensorised Lattice Footplate for a Semi-Active Prosthetic Foot}

\author{Jinze Ge\textsuperscript{1}, Jingcheng Sun\textsuperscript{2} and Chengxu Zhou\textsuperscript{2}
\thanks{This work was partially supported by the Advanced Research and Invention Agency [grant number SMRB-SE01-P06]. 
}
\thanks{\textsuperscript{1}Dept. of Mechanical Engineering, University College London, UK.}
\thanks{\textsuperscript{2}Dept. of Computer Science, University College London, UK. {\tt\small chengxu.zhou@ucl.ac.uk}}
}

\maketitle

\begin{abstract}
This paper investigates whether magnetic plantar sensing can be embedded directly inside the load-bearing compliant element of a low-cost semi-active prosthetic foot. We present a prototype integrating a sensorised 3D-printed lattice footplate, a servo-adjustable hydraulic damper, and a reduced-order ankle model. The damper is experimentally characterised to relate adjustment angle to damping coefficient. Controlled compression tests show tunable lattice stiffness, while cyclic normal loading shows that the embedded sensor tracks the testing-machine reference force, supporting plantar-force estimation without an external insole layer. Static-posture trials under approximately body-weight loading show that forefoot and rearfoot loading distributions are separable across four prescribed stance configurations, providing a preliminary check of the sensing pipeline. A feedforward damping schedule approximates the dorsiflexion trend of a reference ankle trajectory through early-to-mid stance, while exposing the expected limitation that a purely dissipative mechanism cannot generate active push-off. Together, these results demonstrate that sensing can be embedded inside the load-bearing compliant element of a prosthetic foot and used to drive semi-active damping.
\end{abstract}


\section{Introduction}

Lower-limb prosthetic feet are required to provide stable load support, impact attenuation, rollover assistance, and energy return during walking. The need for practical and affordable prosthetic solutions is increasing, as approximately 50,000 people currently live with lower-limb loss in the United Kingdom \cite{ravimaheswaran_2024_time}, while the number of individuals with major lower-limb loss in the United States is projected to increase substantially by 2050 \cite{zieglergraham_2008_estimating}. Among existing prosthetic foot technologies, passive energy-storing-and-returning (ESR) feet remain widely used because of their robustness, low mass, and relatively low cost \cite{versluys_2009_prosthetic,lemoyne_2016_energy}. However, their stiffness and damping characteristics are largely fixed after manufacture and alignment, which limits their ability to adapt to gait phase, walking speed, terrain, and user loading conditions. This can lead to reduced terrain adaptability, limited push-off assistance, compensatory gait strategies, and increased loading on the intact limb \cite{schmalz_2002_energy, au_2009_powered}. In addition, the reliance on carbon-fibre composite structures may increase manufacturing cost and raise sustainability concerns \cite{minuto_2025_design}.

\begin{figure}
    \centering
    \includegraphics[width=0.9\columnwidth]{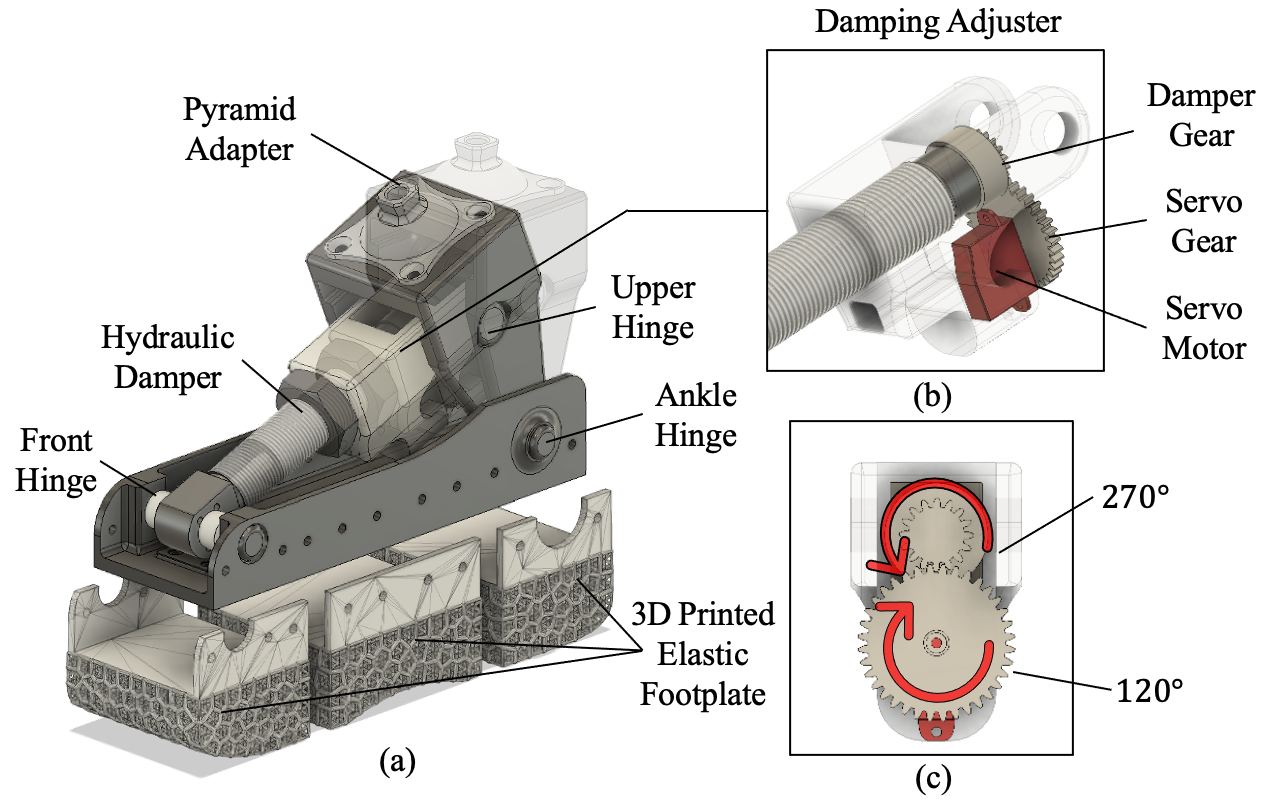}
    \vspace{-2mm}
    \caption{Overview of the proposed semi-active prosthetic foot. 
    (a) CAD assembly showing the articulated ankle mechanism, hydraulic damper, and 3D-printed elastic footplate. 
    (b) Servo-driven damping adjustment module. 
    (c) Gear transmission used to amplify servo rotation for damper adjustment.}
    \label{fig:overview}
    \vspace{-2mm}
\end{figure}

The human ankle--foot complex provides phase-dependent mechanical behaviour during walking, including controlled plantar flexion, controlled dorsiflexion, and powered plantar flexion \cite{versluys_2009_prosthetic, au_2009_powered}. A key feature of the biological ankle is its ability to modulate quasi-stiffness according to walking speed, body mass, and terrain. Powered ankle--foot prostheses have therefore been developed to reproduce biological ankle torque and power more closely, particularly during late stance push-off \cite{au_2009_powered, mazzarini_2023_a}. Nevertheless, the motors, transmissions, batteries, sensors, and embedded control systems required for active power delivery increase device mass, cost, and structural complexity, which may reduce socket comfort and daily clinical practicality \cite{lenzi_2019_design}.

Quasi-passive and semi-active prostheses provide an alternative approach by regulating stiffness, damping, or joint configuration without continuously injecting net positive work. Variable-stiffness and semi-active designs have demonstrated improved ankle motion, energy storage and return, slope adaptation, and reduced contralateral loading in different walking conditions \cite{lapr_2011_simulation, shepherd_2017_the, glanzer_2018_design}. However, many of these systems still rely on specialised mechanisms, costly materials, or limited sensing integration \cite{shepherd_2017_the, glanzer_2018_design}. Beyond ankle-level impedance control, prosthetic foot performance is also strongly affected by the compliance of the footplate and sole. Biomimetic and additively manufactured designs can improve distributed compliance, reduce material usage, and enable patient-specific geometry \cite{pace_2026_the, dhairyakathrotiya_2023_a}. However, most of these structures remain passive and provide limited gait-dependent damping modulation.

Adaptive prosthetic control also requires reliable information about foot--ground interaction. Plantar sensing can support gait-event detection, load estimation, and damping regulation \cite{pandit_2018_an}. However, insole-based sensing remains external to the main load-bearing structure. Recent magnetic tactile sensing approaches show that deformation-induced magnetic-field changes within compliant structures can be used for contact localisation and force estimation \cite{pattabiraman_2025_eflesh}. Embedding this sensing principle directly into a load-bearing prosthetic lattice could hence provide compact plantar force estimation while avoiding a separate external sensing layer.

This paper investigates whether magnetic plantar sensing can be embedded directly into the load-bearing compliant element of a low-cost prosthetic foot, and whether the resulting force readout is informative enough to support a preliminary semi-active damping strategy under controlled prototype conditions. To answer these questions, we design and characterise an integrated prototype, shown in Fig.~\ref{fig:overview}, that combines a sensorised 3D-printed lattice footplate, a servo-adjustable hydraulic damper, and a reduced-order linkage model of the articulated ankle. The main contributions are:
\begin{itemize}
    \item A sensorised lattice footplate that integrates magnetic plantar sensing inside the load-bearing compliant element, eliminating a separate insole sensing layer.

    \item A reduced-order articulated-ankle model coupled with experimental damper characterisation, providing a simulation framework for the proposed mechanism.
    
    \item Mechanical, sensing, and prototype-level characterisation, including a stance-phase simulation against a literature reference and a static-posture preliminary check of the end-to-end sensing pipeline.
\end{itemize}

The remainder of this paper is organised as follows. Section~II describes the system design and modelling framework. Section~III presents the experimental validation, simulation results, and discussion. Section~IV concludes the paper and outlines future work.

\section{System Design and Modelling}

\subsection{Mechanical Architecture of the Proposed Design}

The proposed prosthesis combines an articulated load-bearing frame, an adjustable hydraulic damper, and a 3D-printed elastic footplate, as shown in Fig.~\ref{fig:overview}(a). A standard pyramid adapter connects the device to the prosthetic socket and transfers body weight to the main frame. The upper, ankle, and front hinges constrain motion mainly to the sagittal plane, enabling controlled ankle rotation during stance while maintaining structural support for body-weight transmission.

A hydraulic damper is mounted obliquely between the upper frame and the anterior distal linkage. In the intended stance-phase operation, ankle rotation changes the damper length through the linkage, generating a velocity-dependent resisting force that can regulate the angular response of the ankle. To enable adjustable damping for semi-active modulation, the damper is coupled to a servo-driven gear transmission. As illustrated in Fig.~\ref{fig:overview}(b) and (c), approximately $120^\circ$ of servo rotation is converted into approximately $270^\circ$ of damper adjustment, increasing the available damping range while keeping the actuation module compact.

The distal footplate is fabricated as a compliant 3D-printed lattice rather than a conventional monolithic carbon-fibre spring. This structure provides distributed deformation under plantar loading, contributing to shock attenuation, load redistribution, and controlled rollover during stance. The lattice geometry also allows the footplate stiffness to be tuned through unit-cell size and material selection.

In the overall design, the articulated damper modulates joint-level resistance, whereas the elastic lattice footplate provides localised foot--ground compliance. By combining
these functions, the proposed architecture integrates load-bearing support, adjustable ankle damping, and compliant plantar interaction within a compact semi-active prosthetic-foot platform.

\subsection{Plantar Sensing Module}

\begin{figure}
    \centering
    \includegraphics[width=0.8\columnwidth]{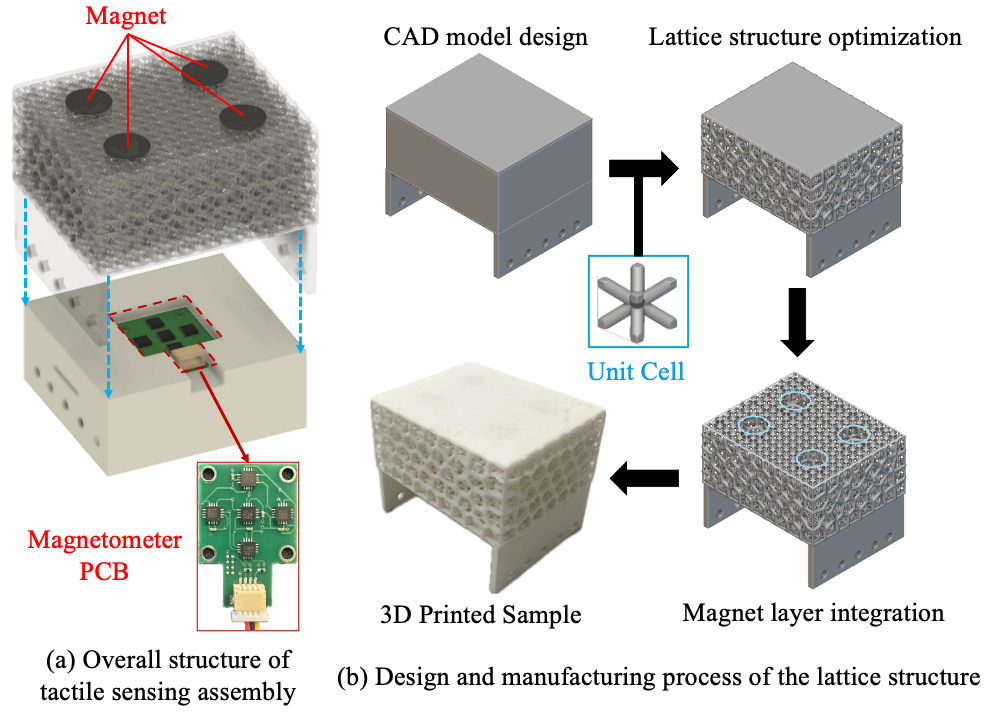}\vspace{-2mm}
    \caption{Embedded plantar sensing module. 
(a) Assembly of the lattice sensing body, embedded magnets, and magnetometer PCB. 
(b) Design and fabrication workflow showing lattice generation, magnet-layer integration, and the PEBA 3D-printed sensing sample.}
    \label{fig:tactile_module}
    \vspace{-3mm}
\end{figure}

The plantar sensing module was inspired by the eFlesh magnetic tactile sensing concept~\cite{pattabiraman_2025_eflesh} and adapted for integration within the 3D-printed lattice footplate. As illustrated in Fig.~\ref{fig:tactile_module}(a), the module consists of a compliant lattice body with embedded permanent magnets positioned above a magnetometer printed circuit board (PCB). Under plantar compression, lattice deformation changes the relative displacement between the magnets and sensor array, producing magnetic-field variations that can be used to estimate contact force and loading distribution.

Fig.~\ref{fig:tactile_module}(b) further shows the sensing body uses a body-centred cubic (BCC) lattice to provide controlled compliance while maintaining structural support. Compared with the original eFlesh structure, the design was adapted for the geometry and load-bearing requirements of the prosthetic foot. Predefined magnet pockets were included in the lattice layer, allowing the print to be paused for magnet insertion and then resumed to encapsulate the magnets. The PCB was mounted beneath the lattice body to complete the sensing assembly. This integrated design enables compact plantar force estimation for static stance-state classification and semi-active damping control. The novelty here lies in the integration of the magnetic sensing principle into a load-bearing prosthetic structure rather than in the magnetic transduction principle itself, which follows~\cite{pattabiraman_2025_eflesh}.

\subsection{Magnetic Plantar Sensing and Force Estimation}

The magnetic sensing signals were processed using a multilayer perceptron (MLP) to estimate plantar contact force from deformation-induced magnetic-field changes. Five three-axis magnetometers provide a 15-dimensional input vector, consisting of the $x$, $y$, and $z$ magnetic-field components from each sensing location. Before inference, the input features were standardised using the mean and standard deviation of the training dataset.

The training data were collected from a controlled Instron normal-compression sequence. A calibrated normal force was applied to the eFlesh module over a force range of 0--700~N, with the reference force increased in 25~N increments. For each force level, approximately 200 magnetic-field samples were recorded, resulting in a dataset of 5600 labelled samples with 15 input features. The dataset was randomly divided into training, validation, and test subsets using a 70\%/15\%/15\% split. The cyclic-compression result shown in Fig.~\ref{fig:tactile_test}(b) was obtained from an independent loading test and was used to evaluate the trained model under time-varying loading and unloading conditions.

The MLP maps the normalised magnetic input to a scalar force estimate, $\hat{F}$. The network contains two fully connected hidden layers with 64 and 32 ReLU-activated nodes and a single force-regression output. The model was trained for 200 epochs using the Adam optimiser with a learning rate of $1\times10^{-3}$ and a mean-squared-error loss function. After training, the final validation MSE was 23.5~N$^2$, corresponding to an RMSE of approximately 4.85~N. The held-out test set produced a mean absolute error of 3.79~N. These results indicate accurate force estimation under controlled normal compression, supporting the present static-posture trial but not yet validating gait-phase classification during walking.

\subsection{Dynamic Modelling of the Articulated Ankle}

\begin{figure}
    \centering
    \includegraphics[width=0.56\columnwidth]{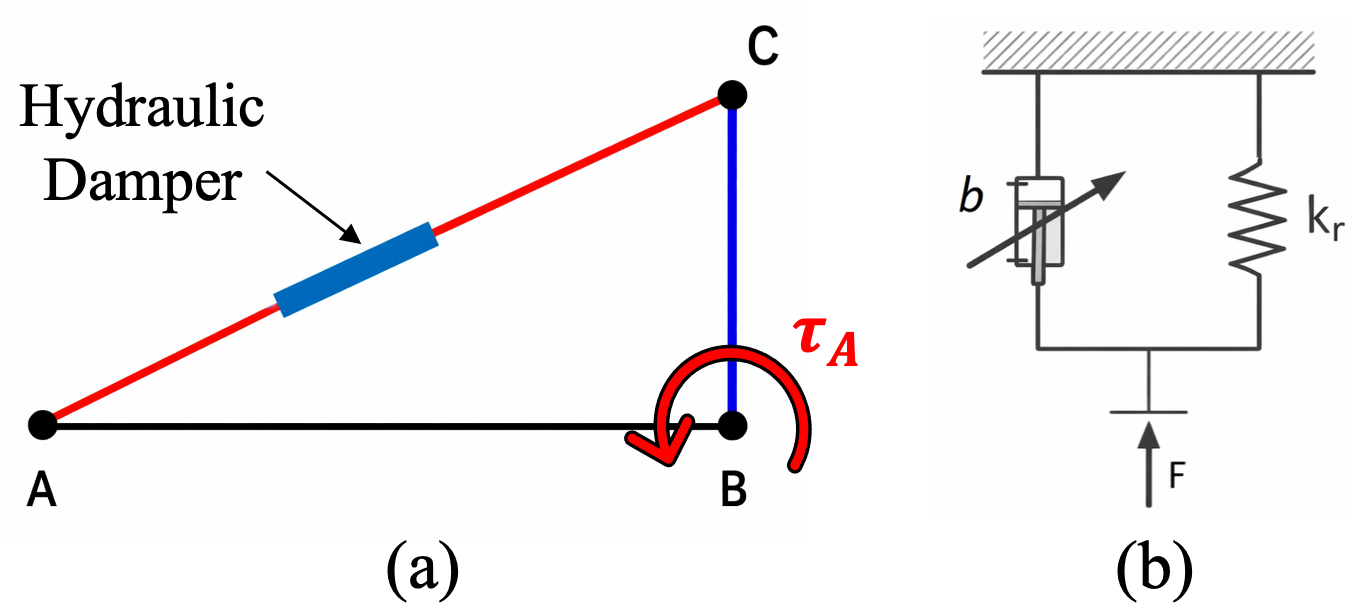}
    \caption{Reduced-order dynamic model of the articulated ankle. 
    (a) Planar linkage model with the hydraulic damper acting along $AC$ and external ankle torque $\tau_A$ applied at joint $B$. 
    (b) Equivalent axial spring--damper model with variable damping coefficient $b$ and return spring stiffness $k_r$.}
    \label{fig:ankle_model}
    \vspace{-5mm}
\end{figure}

A reduced-order sagittal-plane model was developed to describe the ankle response of the articulated mechanism. As shown in Fig.~\ref{fig:ankle_model}(a), the prosthetic ankle is simplified as two rigid links, $AB$ and $BC$, connected at the ankle joint $B$, while the hydraulic damper acts along the line $AC$. The equivalent axial element is represented by a variable damper $b$ in parallel with a return spring $k_r$, as shown in Fig.~\ref{fig:ankle_model}(b).

Let $a=|AB|$ and $h=|BC|$. For an ankle angle $\theta$, the instantaneous damper length is
\begin{equation}
L(\theta)=\sqrt{a^2+h^2-2ah\sin\theta},
\end{equation}
and the reference length at $\theta=0$ is
\begin{equation}
L_0=\sqrt{a^2+h^2}.
\end{equation}
The axial compression of the spring--damper unit is then obtained as
\begin{equation}
x(\theta)=L_0-L(\theta)
\end{equation}
where positive $x$ denotes compression. Differentiating $x(\theta)$ gives the relationship between ankle angular velocity and axial compression velocity:
\begin{equation}
\dot{x}=g(\theta)\dot{\theta}, 
\qquad
g(\theta)=\frac{ah\cos\theta}{L(\theta)}
\end{equation}
where $g(\theta)$ is the geometry-dependent transmission factor between axial force and ankle torque. The spring-generated torque and damping-generated torque are expressed as
\begin{equation}
\tau_s(\theta)
=
-g(\theta)\left[k_r x(\theta)+F_0\right]
\end{equation}
\begin{equation}
\tau_d(\theta,\dot{\theta})
=
\begin{cases}
-b\,g^2(\theta)\dot{\theta} & \dot{x}>0\\
0 & \dot{x}\leq0
\end{cases}
\end{equation}
where $k_r$ is the return-spring stiffness, $F_0$ is the preload force, and $b$ is the variable damping coefficient. The one-way damping term assumes that the hydraulic damper mainly resists compression. The total resisting torque is therefore
\begin{equation}
\tau_{sd}(\theta,\dot{\theta})
=
\tau_s(\theta)+\tau_d(\theta,\dot{\theta})
\end{equation}
with the external ankle torque $\tau_A(t)$ applied at joint $B$, the rotational dynamics are written as
\begin{equation}
J\ddot{\theta}
=
\tau_A(t)+\tau_{sd}(\theta,\dot{\theta})
\end{equation}
where $J$ is the equivalent ankle-joint inertia. This compact model was used to simulate the stance-phase ankle response under gait-derived torque input.

The damping coefficient $b$ was identified from compression tests of the adjustable hydraulic damper. A low-speed quasi-static test first gave $F_0=20~\mathrm{N}$ and $k_r=2000~\mathrm{N/m}$. The damper was then tested at different adjustment angles from $0^\circ$ to $270^\circ$, and the damping coefficient was obtained after subtracting the elastic and preload contributions from the measured force. Following the nonlinear behaviour of valve-controlled hydraulic dampers \cite{xie_2022_research, farahpour_2022_development}, the measured data were fitted as
\begin{equation}
b = c(\phi) = 74.41 \mathrm{e}^{0.02556\phi} + 310.47 
\end{equation}
where $\phi$ is the damper adjustment angle. The fitted model achieved a coefficient of determination of $R^2 = 0.9934$, indicating good agreement with the measured damping trend. This relationship was used to convert the servo-controlled damper angle into the damping coefficient for simulation.

\subsection{Components and Control System}

\begin{figure}
    \centering
    \includegraphics[width=0.8\columnwidth]{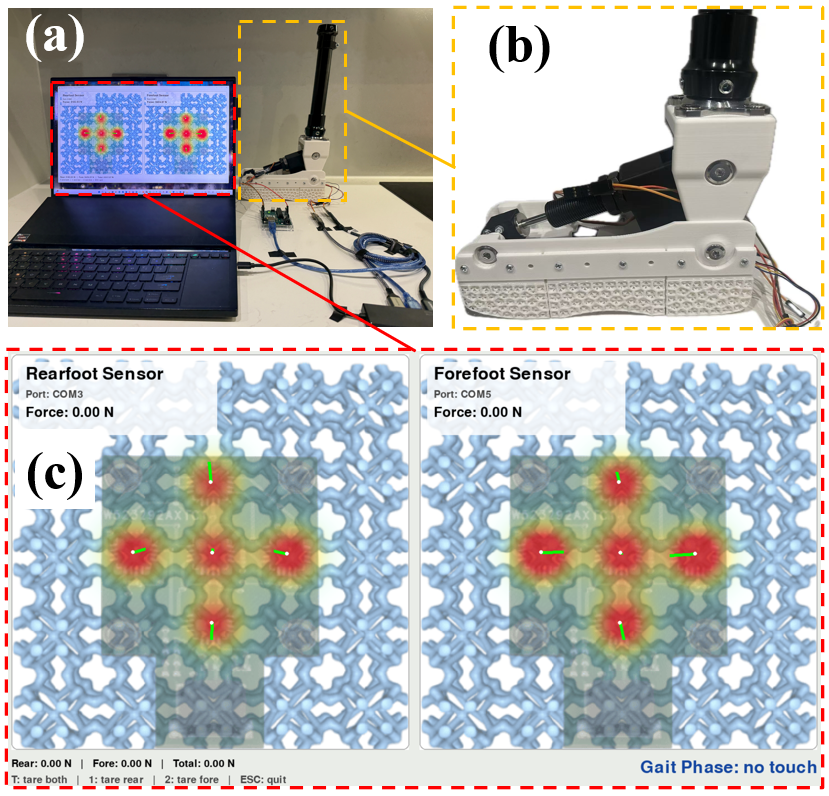}\vspace{-2mm}
    \caption{Prototype sensing and control platform. 
    (a) Experimental setup with host PC, control electronics, and prosthetic foot prototype. 
    (b) Integrated semi-active prosthetic foot with articulated ankle mechanism and lattice footplate. 
    (c) Real-time plantar force visualisation interface showing rearfoot and forefoot force distributions and estimated gait phase.}
    \label{fig:prototype_control}
    \vspace{-2mm}
\end{figure}

The prototype platform integrates embedded plantar sensing, real-time force estimation and visualisation,  stance-state labelling, and servo-based damper adjustment.

As shown in Fig.~\ref{fig:prototype_control}, two magnetic plantar sensing modules are embedded at the rearfoot and forefoot regions of the lattice footplate. The magnetometer signals are streamed to a host PC through microcontroller-based USB communication, where the trained MLP estimates the local plantar forces and displays the load distribution in real time.

Based on the estimated rearfoot and forefoot forces, a preliminary rule-based logic assigns the current loading pattern to representative stance-like configurations, including heel-strike, foot-flat, dorsiflexion, toe-off, and no-contact conditions. The assigned stance-like configuration can then be mapped to a desired damping state of the prosthetic ankle. This command is transmitted to an Arduino-based actuation module, which drives the KST X06H V6.0 micro servo motor connected to the damper gear mechanism. Through the gear transmission, the servo rotation adjusts the hydraulic damper setting and therefore changes the ankle damping coefficient.

This architecture provides a hardware and software pathway toward closed-loop semi-active control, but continuous closed-loop gait control is not validated in this study. Plantar loading is measured by the embedded sensors, processed into force and stance-like state information by the host PC, and used to generate damper-adjustment commands for the actuation module. The adjusted damper can modulate the ankle response in the reduced-order model. In this way, the prototype provides a compact research platform for investigating gait-dependent ankle damping modulation using integrated plantar sensing.

\section{Experimental Validation, Simulation, and Discussion}

The prototype evaluation is organised into three studies that follow the contributions stated in Section~I. Experiment~1 (Section~III-A) characterises the lattice footplate and the embedded sensing module independently under controlled compression, addressing whether sensing can be embedded inside the load-bearing structure without compromising compliance. Experiment~2 (Section~III-B) tests the assembled prototype under prescribed stance postures to verify that the end-to-end sensing pipeline produces separable loading signatures. Simulation~1 (Section~III-C) then evaluates the predicted stance-phase ankle response of the proposed adjustable-damping mechanism against a literature trajectory.

\subsection{Experiment 1---Lattice and Embedded-Sensor Characterisation}

\begin{figure}
    \centering
    \includegraphics[width=0.736\columnwidth]{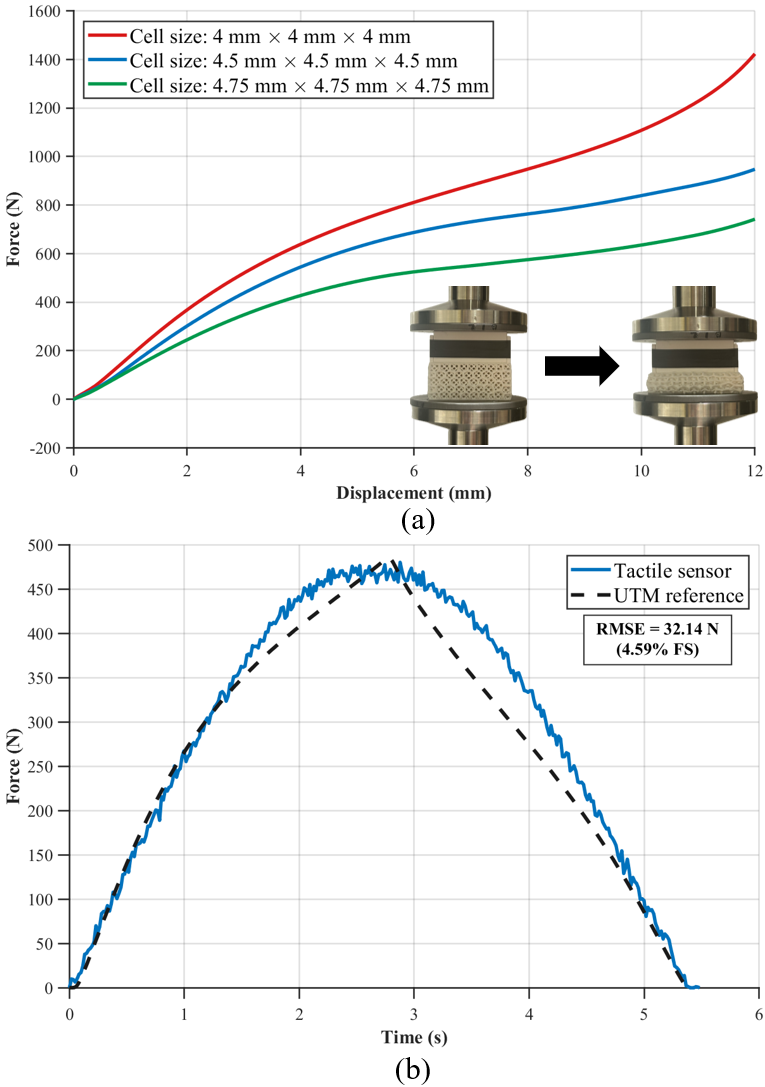}
    \vspace{-2mm}
    \caption{Results for Experiment 1: 
    (a) Force--displacement curves of different lattice unit-cell sizes compared. 
    (b) Embedded sensor force estimation compared with the UTM reference under cyclic compression.}
    \label{fig:tactile_test}
    \vspace{-3mm}
\end{figure}

\subsubsection{Protocol}

Instron compression tests were conducted to evaluate the mechanical response of the 3D-printed lattice footplate and the output of the embedded sensing module. Three lattice unit-cell sizes (4~mm, 4.5~mm, and 4.75~mm) were tested under vertical compression to characterise their force--displacement behaviour. After selecting the lattice geometry, cyclic compression was applied to the assembled sensing module and the sensor-estimated force was compared with the universal testing machine (UTM) measurement.

\subsubsection{Result}

\textit{Lattice tunability.} As shown in  Fig.~\ref{fig:tactile_test} (a), the lattice force--displacement response varies monotonically with unit-cell size: the 4~mm lattice is the stiffest, followed by the 4.5~mm and 4.75~mm cases. This confirms that footplate stiffness can be tuned through unit-cell geometry, and the 4.75~mm lattice was selected as the most compliant option supporting the expected loading range.

\textit{Sensing fidelity inside the load-bearing structure.}
Fig.~\ref{fig:tactile_test} (b) shows the embedded sensor output under cyclic compression. The estimate tracks the UTM reference during loading and unloading, reaching a similar peak of approximately 470~N with an RMSE of 32.14~N, corresponding to 4.59\% full scale. The main deviation occurs near the peak and early unloading. This indicates that plantar force can be estimated from within the load-bearing lattice structure without an external sensing layer under controlled normal-loading conditions, addressing the central question of the paper.

\subsection{Experiment 2---Static-Posture Sensing Trial}

\begin{figure}
    \centering
    \includegraphics[width=0.9\columnwidth]{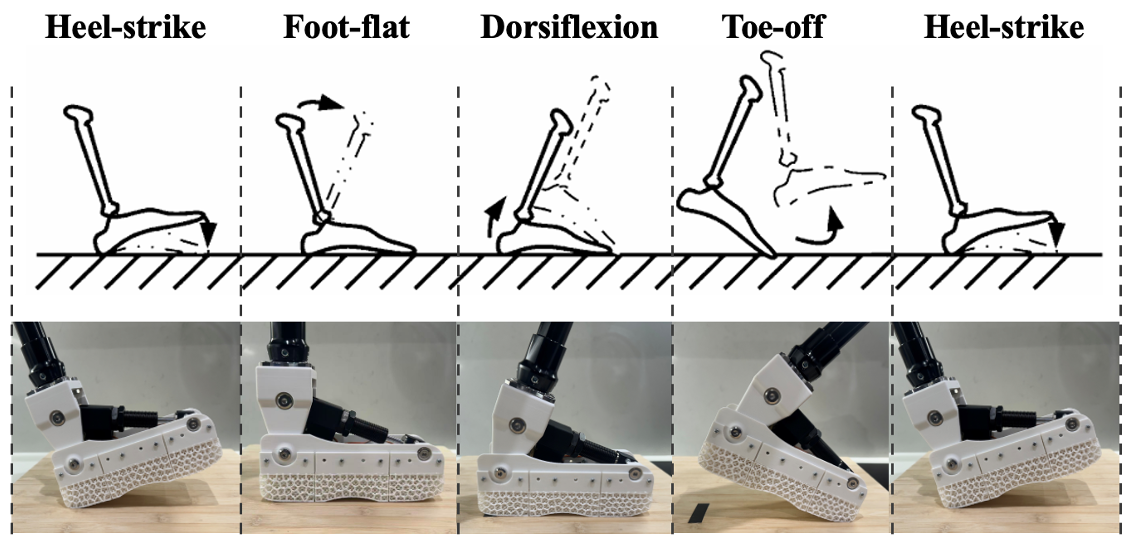}\vspace{-2mm}
    \caption{Results for Experiment 2: prototype evaluation under four representative stance postures, showing the imposed configurations and corresponding prosthetic-foot poses.}
    \label{fig:gait_phase_test}
\end{figure}

\begin{table}
\centering
\caption{Sensed force distributions under four prescribed stance postures, with the rule-based classifier output. Both the postures and the classifier rule were specified a priori; the agreement is therefore a preliminary check of the sensing pipeline rather than a validation of phase detection.}
\label{tab:gait_phase_force}
\resizebox{\columnwidth}{!}{
\begin{tabular}{ccccc}
\hline
\textbf{Imposed posture} & 
\textbf{Forefoot} & 
\textbf{Rearfoot} & 
\textbf{Total} & 
\textbf{Classifier output} \\
 & \textbf{[N]} & \textbf{[N]} & \textbf{[N]} &  \\
\hline
Heel-strike  & 0.00   & 597.88 & 597.88 & Heel-strike \\
Foot-flat    & 259.06 & 346.09 & 605.15 & Foot-flat \\
Dorsiflexion & 319.88 & 288.90 & 608.78 & Dorsiflexion \\
Toe-off      & 626.78 & 0.00   & 626.78 & Toe-off \\
\hline
\end{tabular}
}
\vspace{-2mm}
\end{table}

\subsubsection{Protocol}

The prototype was manually held in four representative stance postures (heel-strike, foot-flat, dorsiflexion, toe-off) under approximately body-weight loading, and the plantar force distribution measured by the forefoot and rearfoot sensing modules was recorded for each posture. A simple rule-based classifier, defined a priori from the relative forefoot-to-rearfoot ratio, was then used to check the end-to-end sensing pipeline.

\subsubsection{Result}

The four imposed postures produced clearly separated loading signatures, as shown in Fig.~\ref{fig:gait_phase_test} and summarised in Table~\ref{tab:gait_phase_force}. Heel-strike loaded only the rearfoot sensor (597.88~N) with the forefoot unloaded; toe-off loaded only the forefoot (626.78~N) with the rearfoot unloaded. Foot-flat produced bilateral loading dominated by the rearfoot (forefoot 259.06~N, rearfoot 346.09~N), and dorsiflexion produced bilateral loading dominated by the forefoot (forefoot 319.88~N, rearfoot 288.90~N). Total measured loads remained between 597~N and 627~N across postures, consistent with the imposed body-weight loading.

As expected for a rule-based classifier evaluated on its own design conditions, the classifier output matched the imposed posture in all four cases. We emphasise that this is a static-posture preliminary check rather than a validation of phase detection during walking; what it does demonstrate is that the four loading patterns are linearly separable in the forefoot/rearfoot plane and that the end-to-end pipeline (sensor $\rightarrow$ MLP force estimate $\rightarrow$ classifier) executes correctly. Quantitative benchmarking under continuous gait, including phase transitions, timing delay, walking-speed variation, multi-cycle robustness, and noise from off-axis or shear loading, is required and is left to future work.

\subsection{Simulation 1---Stance-Phase Ankle Response}

\begin{figure}
    \centering
    \includegraphics[width=0.85\linewidth]{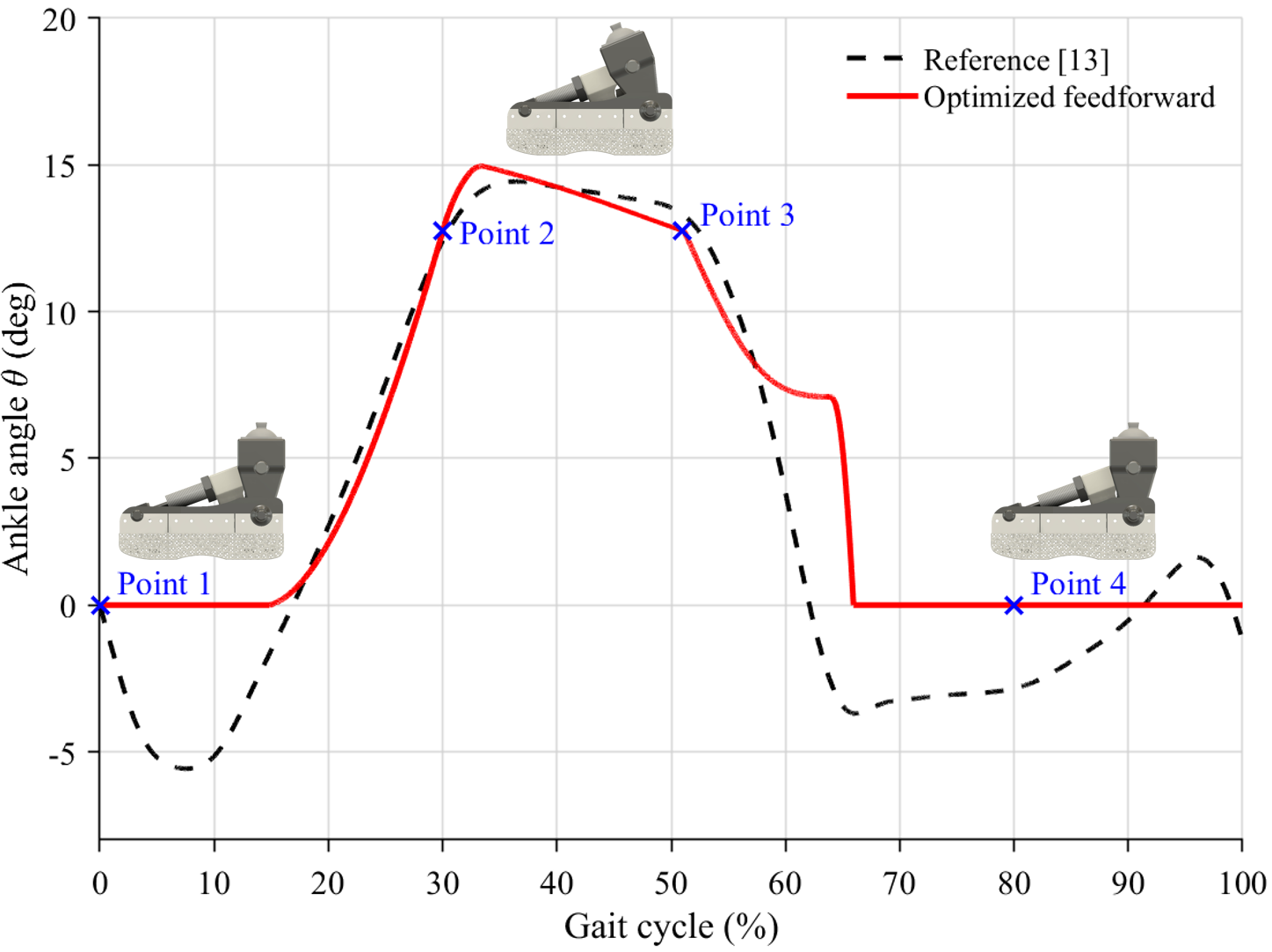}
    \vspace{-2mm}
    \caption{Results for Simulation 1: simulated ankle angle of the proposed prosthesis (red) compared with a reference biological-foot trajectory (black dashed) digitised from~\cite{pace_2026_the}.}
    \label{fig:simu}
    \vspace{-3mm}
\end{figure}

\subsubsection{Protocol}

The sagittal-plane ankle response was simulated using the dynamic model in Section~II-D. A digitised and time-normalised prosthetic ankle-torque profile from~\cite{pace_2026_the} was applied as the input, while the damping coefficient followed a feedforward schedule with four adjustment points over the gait cycle, as shown in Fig.~\ref{fig:simu}: low damping at heel contact, increased damping for dorsiflexion control, high damping for mid-stance stabilisation, and reduced damping during late stance for unloading and reset.

\subsubsection{Result}

The reference trajectory, digitised from~\cite{pace_2026_the}, was used as the target curve because it has been reported to closely match the dorsiflexion trend of a biological foot during stance. As shown in Fig.~\ref{fig:simu}, the simulated ankle angle reproduces the main dorsiflexion trend of this reference: the response rises after early stance, reaches approximately $15^\circ$ at around 35\% of the gait cycle, and remains close to the reference until approximately 55\%. The damping adjustment points regulate the ankle trajectory by increasing resistance during dorsiflexion and reducing damping during late stance. Beyond this phase the simulated response returns rapidly toward $0^\circ$ while the reference continues into plantarflexion before recovery, reflecting the expected limitation of a purely dissipative architecture: the damper can regulate and dissipate energy but cannot generate the active push-off observed in the reference.

\subsection{Discussion}
\subsubsection{Findings against the stated questions}

Two questions were posed in Section~I. First, can magnetic plantar sensing be embedded directly inside the load-bearing compliant element of a low-cost prosthetic foot? The lattice-compression and cyclic-loading results in Section~III-A indicate yes: lattice stiffness can be tuned through unit-cell geometry, and the embedded sensor tracks the UTM reference force inside the load-bearing structure. Second, is the resulting force readout informative enough to support a semi-active damping schedule? The static-posture trial in Section~III-B shows that the four canonical stance configurations produce linearly separable forefoot/rearfoot signatures, and the simulation in Section~III-C shows that, given the characterised damper, a feedforward schedule can track the dorsiflexion phase of a biological-style reference trajectory. 

Together, these results establish the proposed sensing–actuation pairing as a workable basis for semi-active prosthetic-foot research. The most important caveats are: (i) sensing performance has been evaluated only under controlled vertical loading, so the effects of shear forces, uneven contact, material hysteresis, and sensor drift remain to be quantified; (ii) the static-posture classifier is by construction unable to distinguish phase \textit{transitions} during walking, which is the regime that ultimately matters; and (iii) the simulated stance trajectory diverges from the reference in late stance because a purely dissipative architecture cannot reproduce active push-off. The first two are addressed by continuous-gait testing on a treadmill or instrumented walkway; the third is a fundamental limitation of the semi-active concept and not of the prototype.

\subsubsection{Prototype Mass and Cost}

The prototype weighs 0.8~kg in its current 3D-printed form, or approximately 1.0~kg by CAD estimate if the supporting components are replaced with aluminium alloy, remaining close to the estimated mass of a human foot for a 75~kg adult~\cite{fang_2017_anthropometric}. Material and fabrication cost is approximately \$200, against \$1{,}000 to over \$10{,}000 for commercial prosthetic feet~\cite{dhairyakathrotiya_2023_a}. We note that this comparison is between an uncertified research prototype and certified clinical devices, so the cost advantage applies to research-platform use rather than to a clinically deployable product; certified materials, durability testing, and regulatory work would substantially increase the latter.

\section{Conclusion and Future Work}

This paper asked whether plantar sensing can be embedded inside the load-bearing compliant element of a low-cost prosthetic foot and whether the resulting force readout is informative enough to support a semi-active damping schedule under prototype conditions. To address these questions we presented an integrated prototype combining a sensorised 3D-printed lattice footplate, a servo-adjustable hydraulic damper, and a reduced-order articulated-ankle model.

Mechanical characterisation showed that the lattice footplate stiffness is tunable through unit-cell geometry and that the embedded sensor tracks the testing-machine reference force under cyclic compression, supporting the sensing-in-structure claim. The static-posture trial showed that forefoot and rearfoot loading distributions are linearly separable across four representative stance configurations, providing initial evidence that the sensing readout is informative enough for stance-state classification. The reduced-order simulation reproduced the dorsiflexion phase of a reference ankle trajectory from~\cite{pace_2026_the} while confirming that the purely dissipative architecture cannot provide active push-off.

Future work will focus on continuous human gait testing, real-time closed-loop damping control using sensor feedback, fixed-damping baseline comparisons, and durability validation under cyclic, impact, shear, and off-axis loading. Structural redesign of the ankle linkage, lattice stiffness distribution, and damper configuration will also be explored to improve late-stance energy return. Quantitative classifier benchmarking under gait transitions, different walking speeds, repeated gait cycles, timing-delay constraints, and off-axis or shear loading, together with additional sensor calibration and long-term drift assessment, is required before any clinical translation can be considered.

\bibliographystyle{IEEEtran}
\bibliography{reference}

@IEEEtranBSTCTL{IEEEexample:BSTcontrol,
  CTLuse_forced_etal       = "yes",
  CTLmax_names_forced_etal = "1",
  CTLnames_show_etal       = "1"
}

@article{ravimaheswaran_2024_time,
  author = {Ravi Maheswaran and Tong, Thaison and Michaels, Jonathan and Brindley, Paul and Walters, Stephen and Nawaz, Shah},
  publisher = {Wiley},
  title = {Time trends and geographical variation in major lower limb amputation related to peripheral arterial disease in England},
  doi = {10.1093/bjsopen/zrad140},
  volume = {8},
  year = {2024},
  journal = {BJS open}
}

@article{zieglergraham_2008_estimating,
  author = {Ziegler-Graham, Kathryn and MacKenzie, Ellen J. and Ephraim, Patti L. and Travison, Thomas G. and Brookmeyer, Ron},
  pages = {422-429},
  title = {Estimating the Prevalence of Limb Loss in the United States: 2005 to 2050},
  doi = {10.1016/j.apmr.2007.11.005},
  volume = {89},
  year = {2008},
  journal = {Archives of Physical Medicine and Rehabilitation}
}

@article{versluys_2009_prosthetic,
  author = {Versluys, Rino and Beyl, Pieter and Van Damme, Michael and Desomer, Anja and Van Ham, Ronald and Lefeber, Dirk},
  pages = {65-75},
  title = {Prosthetic feet: State-of-the-art review and the importance of mimicking human ankle–foot biomechanics},
  doi = {10.1080/17483100802715092},
  urldate = {2019-11-17},
  volume = {4},
  year = {2009},
  journal = {Disability and Rehabilitation: Assistive Technology}
}

@article{lemoyne_2016_energy,
  author = {LeMoyne, Robert},
  pages = {69-76},
  publisher = {Springer Japan},
  title = {Energy Storage and Return (ESAR) Prosthesis},
  doi = {10.1007/978-4-431-55816-3_6},
  year = {2016},
  journal = {Advances for Prosthetic Technology}
}

@article{schmalz_2002_energy,
  author = {Schmalz, Thomas and Blumentritt, Siegmar and Jarasch, Rolf},
  pages = {255-263},
  title = {Energy expenditure and biomechanical characteristics of lower limb amputee gait: The influence of prosthetic alignment and different prosthetic components},
  doi = {10.1016/s0966-6362(02)00008-5},
  volume = {16},
  year = {2002},
  journal = {Gait \& Posture}
}

@article{au_2009_powered,
  author = {Au, S.K. and Weber, J. and Herr, H.},
  pages = {51-66},
  title = {Powered Ankle--Foot Prosthesis Improves Walking Metabolic Economy},
  doi = {10.1109/tro.2008.2008747},
  urldate = {2020-05-17},
  volume = {25},
  year = {2009},
  journal = {IEEE Transactions on Robotics}
}

@article{minuto_2025_design,
  author = {Minuto, Matilde and Emanuele Gruppioni and Mattia Frascio},
  publisher = {SAGE},
  title = {Design and optimization of bamboo-laminate prosthetic feet: A sustainable, cost-effective alternative to carbon fiber models},
  doi = {10.1177/14644207251339669},
  year = {2025},
  journal = {Proceedings of the Institution of Mechanical Engineers Part L Journal of Materials Design and Applications}
}

@article{mazzarini_2023_a,
  author = {Mazzarini, Alessandro and Fantozzi, Matteo and Papapicco, Vito and Ilaria Fagioli and Lanotte, Francesco and Baldoni, Andrea and Filippo Dell'Agnello and Ferrara, Paolo and Tommaso Ciapetti and Raffaele Molino Lova and Emanuele Gruppioni and Trigili, Emilio and Crea, Simona and Vitiello, Nicola},
  title = {A low-power ankle-foot prosthesis for push-off enhancement},
  doi = {10.1017/wtc.2023.13},
  volume = {4},
  year = {2023},
  journal = {Wearable technologies}
}

@article{lenzi_2019_design,
  author = {Lenzi, Tommaso and Cempini, Marco and Hargrove, Levi J. and Kuiken, Todd A.},
  pages = {471-482},
  title = {Design, Development, and Validation of a Lightweight Nonbackdrivable Robotic Ankle Prosthesis},
  doi = {10.1109/tmech.2019.2892609},
  volume = {24},
  year = {2019},
  journal = {IEEE/ASME Transactions on Mechatronics}
}

@article{shepherd_2017_the,
  author = {Shepherd, Max K. and Rouse, Elliott J.},
  pages = {2375-2386},
  title = {The {VSPA} Foot: A Quasi-Passive Ankle-Foot Prosthesis With Continuously Variable Stiffness},
  doi = {10.1109/tnsre.2017.2750113},
  volume = {25},
  year = {2017},
  journal = {IEEE Transactions on Neural Systems and Rehabilitation Engineering}
}

@article{glanzer_2018_design,
  author = {Glanzer, Evan M. and Adamczyk, Peter G.},
  pages = {2351-2359},
  title = {Design and Validation of a Semi-Active Variable Stiffness Foot Prosthesis},
  doi = {10.1109/tnsre.2018.2877962},
  urldate = {2022-06-08},
  volume = {26},
  year = {2018},
  journal = {IEEE Transactions on Neural Systems and Rehabilitation Engineering}
}

@article{lapr_2011_simulation,
  author = {LaPrè, Andrew K and Sup, Frank},
  pages = {587-90},
  title = {Simulation of a slope adapting ankle prosthesis provided by semi-active damping},
  doi = {10.1109/IEMBS.2011.6090110},
  volume = {2011},
  year = {2011},
  journal = {International Conference of the IEEE Engineering in Medicine and Biology Society}
}

@article{pace_2026_the,
  author = {Pace, Anna and Dimitrov, Hristo and Jakubowitz, Eike and Piazza, Cristina and Grioli, Giorgio and Proksch, Lukas and Aszmann, Oskar C. and Farina, Dario and Bicchi, Antonio and Catalano, Manuel G.},
  title = {The {SoftFoot Pro}: an anthropomorphic and adaptive soft articulated prosthetic foot},
  doi = {10.1038/s41467-025-68194-2},
  urldate = {2026-01-27},
  year = {2026},
  journal = {Nature Communications}
}

@article{dhairyakathrotiya_2023_a,
  author = {Dhairya Kathrotiya and Yusuf, Abid and Ranjeet Kumar Bhagchandani and Gupta, Satyapriya},
  title = {A Study for the development of prosthetic foot by additive manufacturing},
  doi = {10.1007/s40430-023-04107-y},
  volume = {45},
  year = {2023},
  journal = {Journal of the Brazilian Society of Mechanical Sciences and Engineering}
}

@article{pandit_2018_an,
  author = {Pandit, Srinivas and Godiyal, Anoop Kant and Vimal, Amit Kumar and Singh, Upinderpal and Joshi, Deepak and Kalyanasundaram, Dinesh},
  pages = {706},
  title = {An Affordable Insole-Sensor-Based Trans-Femoral Prosthesis for Normal Gait},
  doi = {10.3390/s18030706},
  volume = {18},
  year = {2018},
  journal = {Sensors}
}

@article{pattabiraman_2025_eflesh,
  author = {Pattabiraman, Venkatesh and Huang, Zizhou and Panozzo, Daniele and Zorin, Denis and Pinto, Lerrel and Bhirangi, Raunaq},
  title = {{eFlesh}: Highly customizable Magnetic Touch Sensing using Cut-Cell Microstructures},
  year = {2025},
  journal={arXiv:2506.09994}
}

@article{xie_2022_research,
  author = {Xie, Fangwei and Shi, Xiuwei and Cao, Jinxin and Ding, Zhiwen and Yu, Chengcheng and Gao, Yonghua},
  publisher = {Frontiers Media},
  title = {Research on Damping Contribution Rate of Key Parameters of Valve-Controlled Damping Adjustable Damper},
  doi = {10.3389/fenrg.2022.854529},
  volume = {10},
  year = {2022},
  journal = {Frontiers in Energy Research}
}

@article{farahpour_2022_development,
  author = {Farahpour, Hengameh and Hejazi, Farzad},
  pages = {1295-1322},
  publisher = {Elsevier},
  title = {Development of adjustable fluid damper device for the bridges subjected to traffic loads},
  doi = {10.1016/j.istruc.2022.11.136},
  volume = {47},
  year = {2022},
  journal = {Structures}
}

@article{fang_2017_anthropometric,
  author = {Fang, Y. and Morse, L.R. and Nguyen, N. and Tsantes, N.G. and Troy, K.L.},
  pages = {11–17},
  title = {Anthropometric and Biomechanical Characteristics of Body Segments in Persons with Spinal Cord Injury},
  doi = {10.1016/j.jbiomech.2017.01.036},
  volume = {55},
  year = {2017},
  journal = {Journal of biomechanics}
}

\end{document}